\def\year{2020}\relax
\documentclass[letterpaper]{article} 
\usepackage{aaai20}  
\usepackage{times}  
\usepackage{helvet} 
\usepackage{courier}  
\usepackage[hyphens]{url}  
\usepackage{graphicx} 
\usepackage{graphicx}  
\usepackage{xspace}
\usepackage{amsmath}
\usepackage{comment}
\usepackage{array}
\usepackage{color}
\usepackage{mathtools}
\usepackage{dsfont}
\usepackage{booktabs} 

\newcommand{\system}{KG-Story\xspace}

\newcommand\blfootnote[1]{%
  \begingroup
  \renewcommand\thefootnote{}\footnote{#1}%
  \addtocounter{footnote}{-1}%
  \endgroup
}
\title{Knowledge-Enriched Visual Storytelling}
\author{
    \begin{tabular}[c]{@{}c@{}}
    Chao-Chun Hsu\textsuperscript{1*}~~~
    Zi-Yuan Chen\textsuperscript{2*}~~~
    Chi-Yang Hsu\textsuperscript{3}\\
    Chih-Chia Li\textsuperscript{4}~~~
    Tzu-Yuan Lin\textsuperscript{5}~~~
    Ting-Hao (Kenneth) Huang\textsuperscript{3}~~~
    Lun-Wei Ku\textsuperscript{2,6} \end{tabular}\\
    \textsuperscript{1}{University of Colorado Boulder},~~~
    \textsuperscript{2}{Academia Sinica},~~~
    \textsuperscript{3}{Pennsylvania State University},\\
    \textsuperscript{4}{National Chiao Tung University},~~~
    \textsuperscript{5}{National Taiwan University},\\
    \textsuperscript{6}{Most Joint Research Center for AI Technology and All Vista Healthcare}\\
    chao-chun.hsu@colorado.edu,~~~
    \{zychen, lwku\}@iis.sinica.edu.tw,~~~
    \{cxh5437, txh710\}@psu.edu~~~
}

\begin{document}
\maketitle

\begin{abstract}Stories are diverse and highly personalized, resulting in a large possible
output space for story generation.
Existing end-to-end approaches produce monotonous stories because they are
limited to the vocabulary and knowledge in a single training dataset.
This paper introduces \textbf{\system}, a three-stage framework that allows the
story generation model to take advantage of external \textbf{K}nowledge
\textbf{G}raphs to produce interesting stories.
\system distills a set of representative words from the input
prompts,
enriches the word set by using external knowledge graphs, and finally
generates stories based on the enriched word set.
This \textit{distill-enrich-generate} framework allows the use of external
resources not only for the enrichment phase, but also for the distillation and
generation phases.
In this paper, we show the superiority of \system for visual storytelling,
where the input prompt is a sequence of five photos and the output is a
short story.
Per the human ranking evaluation, stories generated by \system are on average ranked better than that of the state-of-the-art systems.
Our code and output stories are available at \url{https://github.com/zychen423/KE-VIST}.

\end{abstract}

\section{Introduction}
\blfootnote{\\$^*${denotes equal contribution}}

\begin{figure}[t]
    \centering
    \includegraphics[width=\columnwidth]{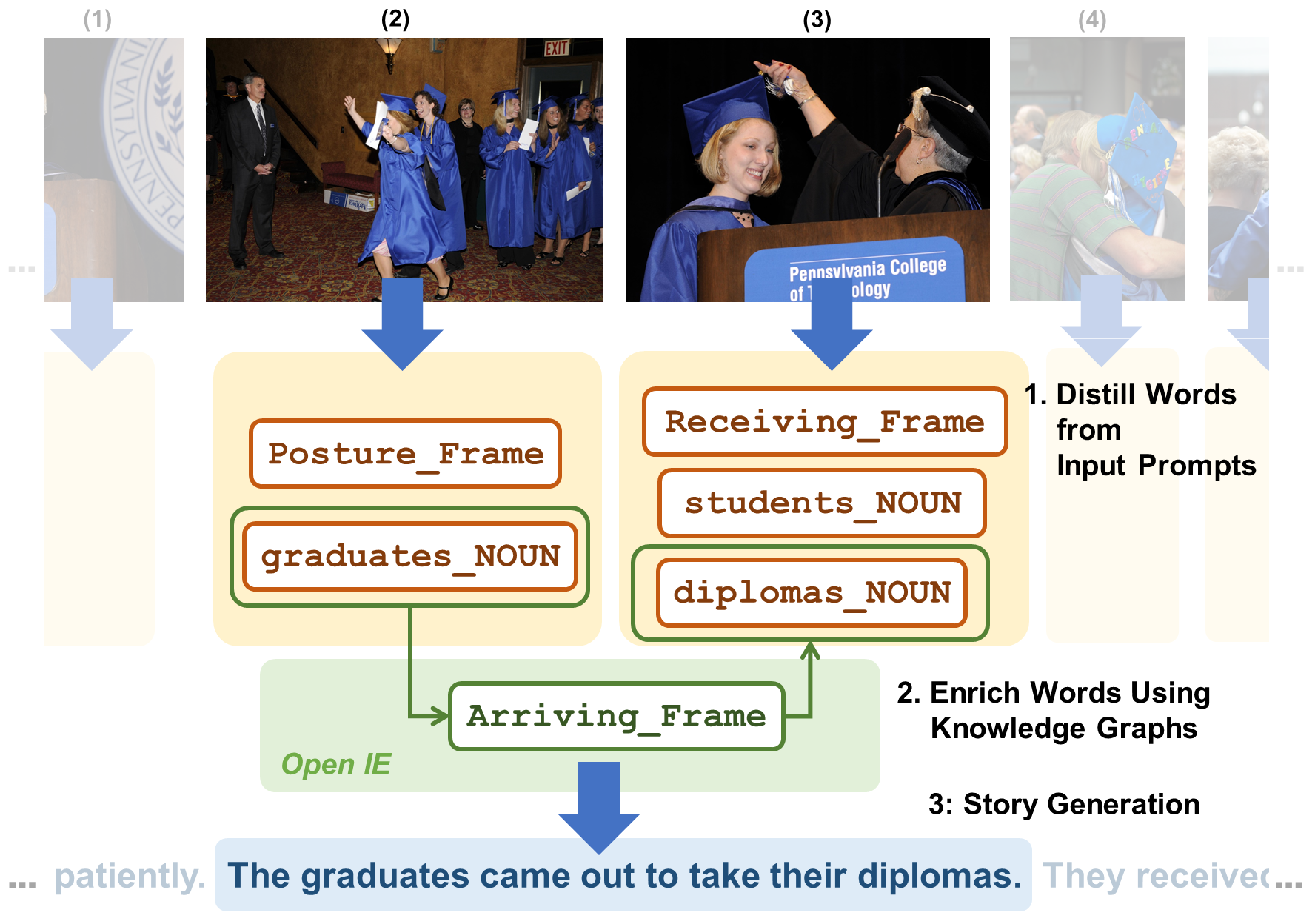}
	 \caption{Overview of \system. \system first {\em (i)} distills a set of representative words from the input prompts and
{\em (ii)} enriches the word set by using external knowledge graphs, and
{\em (iii)} generates stories based on the enriched word set.}
    \label{fig:system}
\end{figure}

Stories are diverse and highly personalized.
Namely, story generation models, whether using text or images as input
prompts, aim at a large possible output space with rich information and
vocabulary.
One classic example is visual storytelling (VIST), an
interdisciplinary task that takes a sequence of five photos as the input and
generates a short story as the output~\cite{huang2016visual}.
However, existing visual storytelling approaches produce monotonous stories
with repetitive text and low lexical diversity~\cite{hsu-etal-2019-visual}.
We believe three major factors contribute to this problem.
First, most prior work uses only a single training set ({\em i.e.},
the VIST dataset) in an end-to-end
manner~\cite{wang2018nometics,Kim2018GLACNG,wang2018show}.
Although this can result in legitimate stories, the end-to-end architecture
makes it difficult to use external data.
The generated stories are thus limited in terms of vocabulary and knowledge in VIST.
Similar phenomena are observed for text story generation models using a single
training set, where the output can be ``fairly generic''~\cite{fan2018hierarchical}.
Second, the VIST dataset is considerably smaller than that of other text
generation tasks.
The MSCOCO dataset for image captioning contains 995,684
captions~\cite{lin2014microsoft};
the VQA dataset for visual question answering contains $\sim$0.76M questions
and $\sim$10M answers~\cite{antol2015vqa}; and the ROC dataset for text story
generation contains 98,159 stories~\cite{mostafazadeh2016roc}.
The VIST dataset, in contrast, contains only 49,603 stories.
Thus it is not surprising that stories generated using VIST show low lexical
diversity.
Finally, no existing work takes into account relations between
photos.
Visual storytelling was introduced as an artificial intelligence task that attempts to
``interpret causal structure'' and ``make sense of visual input to tie
disparate moments together''~\cite{huang2016visual}.
However, existing approaches still treat it as a sequential image captioning
problem and omit relations between images.

We introduce \system, a three-stage framework that allows the
story generation model to take advantage of external resources, especially
knowledge graphs, to produce interesting stories.
\system first distills a set of representative words from the input prompts and
enriches the word set using external knowledge graphs, and finally
generates stories based on the enriched word set.
This \textit{distill-enrich-generate} framework allows the use of external
resources not only for the enrichment phase, but also for the distillation and
generation phases.
Figure~\ref{fig:system} overviews the workflow of \system.
\system first distills a set of words from each image, respectively.
For example, the words {\tt Posture\_Frame} and {\tt graduates\_NOUN} are extracted from image (2), and the words {\tt Receiving\_Frame}, {\tt students\_NOUN} and {\tt diplomas\_NOUN} are extracted from image (3).
In stage 2, \system then searches in a knowledge graph to find potential relations between word pairs across images.
\system uses a scoring function to rate potential relations when multiple relations are found.
In Figure~\ref{fig:system}, the system finds the relation {\tt Arriving\_Frame} between {\tt graduates\_NOUN} and {\tt diplomas\_NOUN}, denoted as $\text{graduates\_NOUN} \xrightarrow{\text{Arriving\_Frame}} \text{diplomas\_NOUN}$.
Finally, in stage 3, \system uses an story generator to produce the final story sentence using all ingredients.
The {\tt Arriving\_Frame} is defined as ``An object Theme moves in the direction of a Goal.'' in the knowledge graph (FrameNet), illustrating a graduates' common body motion in the commencement. 
As a result, the phrase ``came out to take'' in the final output sentence captures this knowledge extracted by \system.

The contributions of this paper are twofold. First, the proposed framework opens up possibilities to using large external data sources to improve
automated visual storytelling, instead of struggling with specialized small
data sources. 
Second, the proposed model leverages external knowledge to explore the relations between images and increases the output text diversity. We show in this paper by human ranking evaluation that the stories generated by \system are on average of higher quality than those from the state-of-the-art systems.

\section{Related Work}
As researchers have found that middle-level abstraction can result in more
coherent stories,~\cite{yao2019plan} show that more diverse stories are
generated by using a pipeline involving first planning the overall storyline
and then composing the story.
They propose an event-based method which transforms the previous story into
event representations, predicts successive events, and generates a story using
the predicted events~\cite{martin2018event}.
Similarly,~\cite{xu2018skeleton} propose a skeleton-based narrative story
generation method which uses a skeleton as a middle-level abstraction layer.
However, 
although multi-stage text story generation has been a focus of recent studies,
limited effort has been expended on multi-stage visual storytelling.
One of the few exceptions is work done
by Huang et al.~\shortcite{DBLP:journals/corr/abs-1805-08191}.
Leveraging the middle-level abstraction as well, they propose a two-level
hierarchical decoder model for the VIST task.
They first predict a topic ({\em e.g.}, indoors, kids, the baby) for each image
in the sequence, after which the low-level decoder generates a sentence for each image
conditioned on the given topic.
The two decoders are jointly trained end-to-end using reinforcement learning.
As the topics can be selected only from fixed clusters using a discriminative
model, the ability of story planning is limited. 
Furthermore, under an end-to-end training process, extra plain text and knowledge
base data cannot be utilized to better enrich the story.
In a similar task,~\cite{Alexander2018Semstyle} propose a two-stage style
transfer model for image captioning.
Their first stage consists of keyword prediction for the caption of each
independent image, and then a stylish caption is generated using an RNN trained
on story corpora.
However, each such caption is a single story-like sentence and is independent
of other captions; combined, the captions do not constitute a context-coherent
story.

\section{\system}

In this work, a three-stage approach is proposed for knowledge-enriched visual
storytelling.

\begin{description}
    \item[Stage 1: Word distillation from input prompts.]
	 Given an image, we train an image-to-term model to distill
	 terms from each of the input images; this can be regarded as word-form conceptual representation.
    \item[Stage 2: Word enrichment using knowledge graphs.]
	 With the five sets of terms extracted from a sequence of images in the previous step, we utilize an external
	 knowledge graph to identify possible links between sets, and generate the
	 final enriched term path for story generation. 
    \item[Stage 3: Story generation.]
	 We use a Transformer architecture to transform term paths into stories. A
	 length-difference positional encoding and a repetition penalty are used in
	 the proposed Transformer; we also apply a novel anaphoric expression
	 generator.
\end{description}

In the following, we describe the three steps in detail.

\subsection{Stage 1: Word Distillation from Input Prompts}

\begin{figure}[t]
    \centering
    \includegraphics[
    width=0.8\columnwidth]{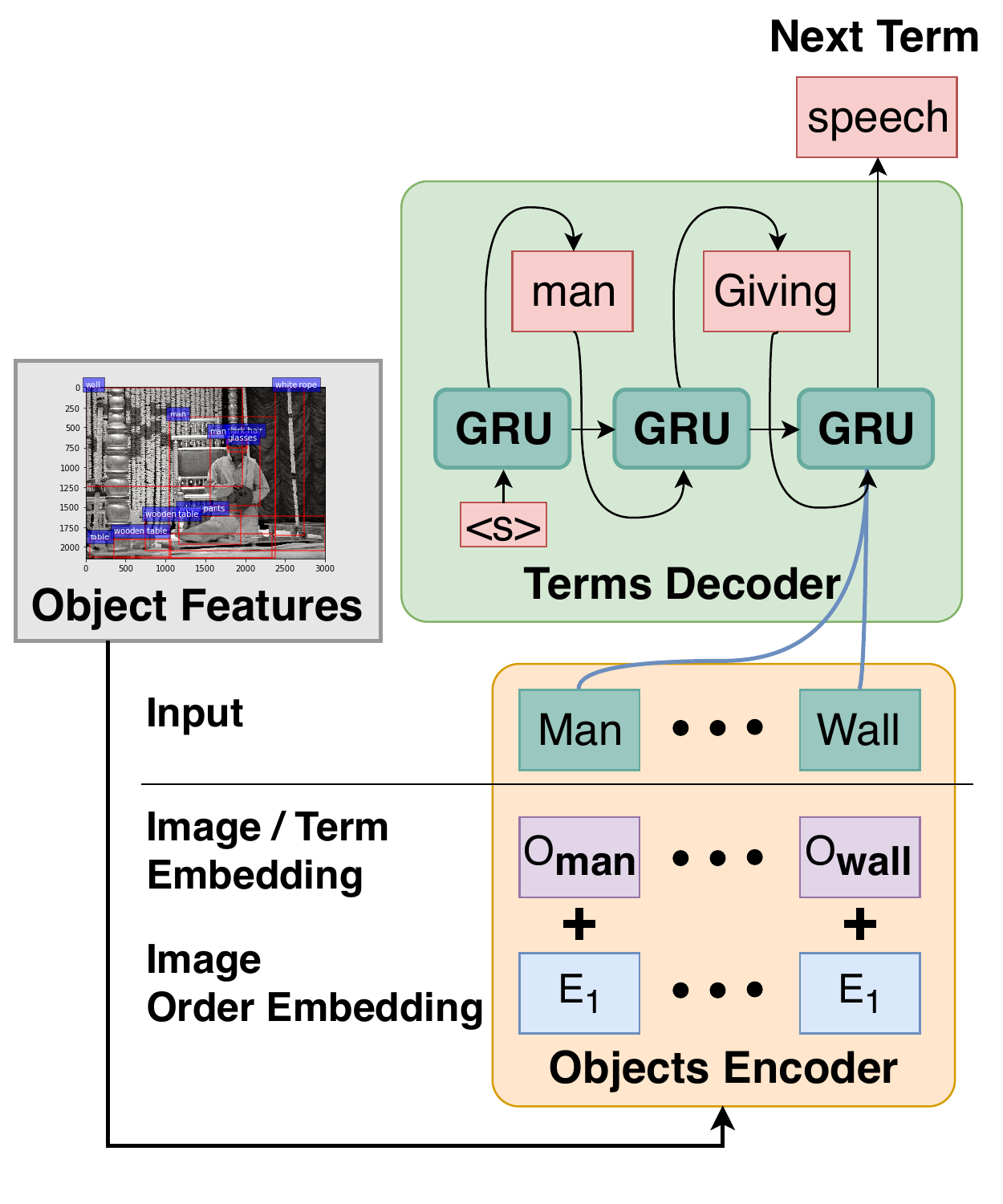}
	 \caption{Stage 1: Image distillation model. In Stage 1, \system extract terms from the image. We build a model that uses a pre-trained Faster R-CNN as the image feature extractor and a Transformer-GRU as the term predictor.}

    \label{fig:step1}
\end{figure}

In Stage 1, we extract terms from the image. To this end, we build a model that uses a pre-trained Faster R-CNN~\cite{ren2015faster,anderson2018bottom} as the image feature extractor and a Transformer-GRU as term predictor.

The Faster R-CNN model was originally trained for the object detection task. Here we extract the features of its predicted object as the image representation. To reduce computational complexity, only the object features within the top 25 confidence scores are used.

As shown in Figure~\ref{fig:step1}, the image representation is fed into a Transformer encoder~\cite{vaswani2017attention} and a GRU~\cite{chung2014empirical} decoder with an attention mechanism~\cite{bahdanau2014neural} as our term prediction model.
The only modification to the Transformer for distillation is the positional encoding.
In contrast to the positional encoding in the original setting, object
features are summed with trainable image-order embeddings as input, as
objects in the same image are sequentially independent:
\begin{align*}
x_i = Wo_i + \text{Order\_Embedding}(t)
\end{align*}
where $o_i$ is the $i$-th object and $t$ is the order of the image.
$W^{2048\times D}$ is a matrix that transforms the 2048-dimensional $o_i$ into
dimension $D$.
Then object features are fed into the term prediction model to generate the
terms for each photo.
During decoding, we use an intra-sentence repetition penalty with beam
search to reduce redundancy in the visual story~\cite{hsu2018using}. 
The term score at each beam search step is computed as
\begin{align*}
\mbox{beam\_score}(x) = \log(p(x))-1e^{19}\cdot\mathds{1}(x{\in}S),
\end{align*}
where $p(x)$ is the model's output probability for each term, the last term
denotes repetition penalty, and set $S$ denotes terms that have already been
predicted in an image.

\subsection{Stage 2: Word Enrichment Using Knowledge Graphs}
Given a sequence of images for story generation, we have observed that nearly 
irrelevant sequential images are common.
In previous end-to-end models, such a phenomenon hurts story generation and causes
models to create caption-like incoherent stories which are relatively boring
and semantically disconnected.
To take this into account while enriching the story, we introduce
semantic terms as the intermediate and link terms in two adjacent images using
the relations provided by the knowledge graph.
In the knowledge graph, real world knowledge is encoded by entities and their
relationship in the form of tuples \emph{(head, rela, tail)}, indicating that
\emph{head} has a \emph{rela} relationship with \emph{tail}. 

More specifically, given the terms of a sequence of image predicted in the previous step $\{m^1_{1},
.., m^t_i, .. ,m^5_{N_5}\}$, where $m^t_i$ denotes the $i$-th term distilled
from the $t$-th image, we pair the terms from two consecutive images and
query the knowledge graph for all possible tuples $\{...., (m^t_i, r_k,
m^{t+1}_j), ....\}$ where $r_k$ is a one-hop relation that links $m^t_i$ and
$m^{t+1}_j$ in the knowledge graph.
Furthermore, we also consider two-hop relations, which can convey indirect
relationships, to enrich the story by adding $(m^t_i, r_k, m_{\mathit{middle}}, r_q,
m^{t+1}_j)$ if such a relation exists in the knowledge graph.

After all tuples of one- and two-hop relations are extracted, we 
construct candidate terms from them and insert each relation into the term set
for story generation, i.e., the extracted $(m^l_i, r_k, m^{l+1}_j)$ or
$(m^l_i, r_k, m_{\mathit{middle}}, r_q, m^{l+1}_j)$, which contains the head term from image
$l$ and the tail term from image ${l+1}$, is inserted between images $l$ and
${l+1}$ to generate an additional story sentence, 
as if these were the terms distilled in between the images.   

With all possible term sets constructed, we must select the most reasonable
one for the next term-to-story generation step.
Hence we train an RNN-based language model on all the available textual stories. 
A language model estimates the probability distribution of a corpus
$$ P(U) = \Sigma \text{logP}(u_i|u_{1},...,u{i-1}) $$
where $U = {u_1, ..., u_n}$ is the corpus.
Here we use it to compute the probability of a term conditioned on the other
terms existing previously.
Thus we obtain the perplexity of this term path, choose that term path with the
lowest perplexity, and feed it to the next step.

Actually, Stage 2 mimics the way people generate a story based on two
irrelevant images, that is, relating two images through imagination. Here the
knowledge graph serves as the source of ideas connecting two images and the
language model then ensures the coherence of the generated story when using the
selected idea.

\subsection{Stage 3: Story Generation}

\begin{figure}[t]
    \centering
    \includegraphics[
    width=1\columnwidth]{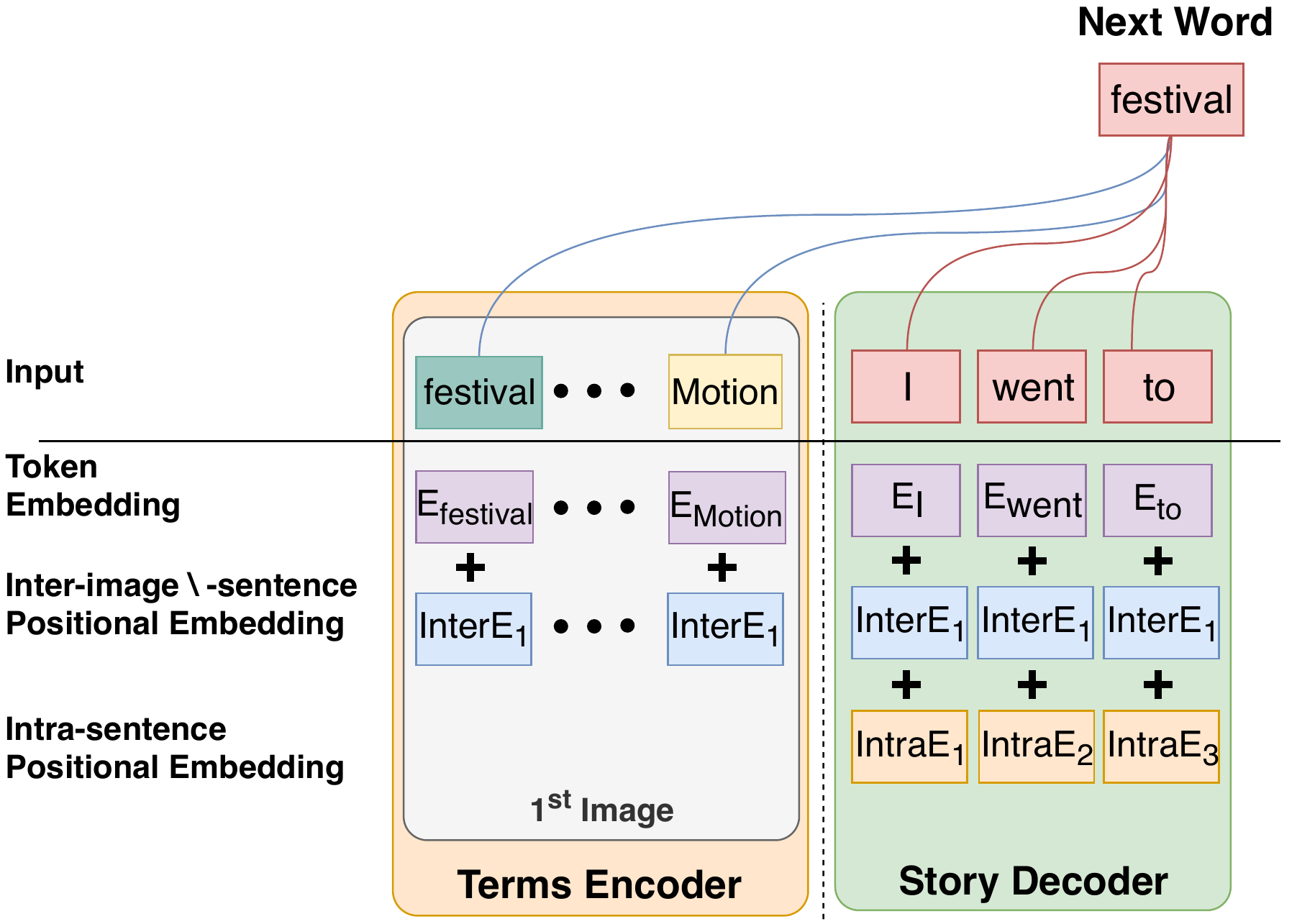}
	 \caption{Stage 3: Story generation model. For the story generation step, we leverage
the Transformer with the input ({\em i.e.}, the term set) from Stage 2. }

    \label{fig:step3}
\end{figure}

For the story generation step, we leverage
the Transformer~\cite{vaswani2017attention} shown in Figure~\ref{fig:step3} with the
input, i.e., the term set, from Stage 2. 
We add to the original Transformer model three different modifications:
{\em (i)} length-difference positional encoding for variable-length story generation,
{\em (ii)} anaphoric expression generation for the unification of anaphor representation,
and {\em (iii)} a repetition penalty for removing redundancy.

\paragraph{Length-Difference Positional Encoding}
The sinusoidal positional encoding method~\cite{vaswani2017attention} inserts
the absolute positions to a sinusoidal function to create a positional
embedding. However, when the model is used to generate variable-length stories,
this positional encoding method makes it difficult for the model to recognize
sentence position in a story. For the visual storytelling task, all the
samples contain five images, each of which is described generally in one sentence
for the corresponding storyline; thus the generated stories always contain five sentences, which
prevents us from adding additional ingredients to enrich them. To tackle this
problem, we adopt length-difference positional encoding
(LDPE)~\cite{Takase_2019} as shown in Equations~(\ref{eqn:LDPE1}) and
(\ref{eqn:LDPE2}), where $\mathit{pos}$ is the position, $d$ is the embedding
size, $\mathit{len}$ is the length constraint, $i$ is the dimension of the sinusoidal
positional encoding, and ${10000^{2i/d}} \times {2}$ is the period of a
sinusoidal function. 
LDPE allows the Transformer to learn positional embeddings with variable-length stories.
Unlike positional encoding that uses absolute positions, LDPE
returns the remaining length of a story. As a result, the terminal position is
identical for dissimilar story lengths. Consequently, the model learns to
generate ending sentence for stories with different lengths.
\begin{equation}
\label{eqn:LDPE1}
\mathit{LDPE}_{(\mathit{pos},\mathit{len},2i)} = \sin \biggl(\frac{\mathit{len}-\mathit{pos}}{10000^{\frac{2i}{d}}}\biggr)
\end{equation}

\begin{equation}
\label{eqn:LDPE2}
\mathit{LDPE}_{(\mathit{pos},\mathit{len},2i+1)} = \cos \biggl(\frac{\mathit{len}-\mathit{pos}}{10000^{\frac{2i}{d}}}\biggr)
\end{equation}

\paragraph{Anaphoric Expression Generation}
To enable the use of pronouns in the enhanced \system model for anaphoric
expression generation, we adopt a coreference replacement strategy.
To replace coreferences, we first use a coreference
resolution tool\footnote{NeuralCoref 4.0: Coreference Resolution in spaCy with Neural Networks. https://github.com/huggingface/neuralcoref}
on the stories to find the original mention of each pronoun and replace the
pronouns with their root entities. The entity types are the ten defined in the
OntoNotes dataset~\cite{weischedel2013ontonotes}, including PERSON,
PRODUCT, ORG, and so on. Open-SESAME then extracts terms again from the 
stories in which the coreferences have been replaced; 
these extracted terms are then used as the input~-- 
with the original stories as the output~-- to train the story generator
in Stage 2. As such, the story generator learns to generate pronouns given
two successive identical noun terms. Table~\ref{tab:term_example} shows a
short visual story of two successive images where both contain the same
object (a dog), and the corresponding gold story reads ``\emph{The dog is ready to
go. He is playing on the ground.}'' In this example, the term predictor 
predicts \emph{dog} for both images, but the term extractor cannot extract
\emph{dog} from the second gold sentence. To bridge this gap, coreference
replacement replaces the pronoun with its original mention ``\emph{the dog}'' in the second
gold sentence, which enables the story generator to learn to generate anaphoric
expressions when seeing multiple mentions of ``\emph{the dog}''.

\begin{table}[]
\resizebox{1\columnwidth}{!}{
\begin{tabular}{l|l|l}
 &\includegraphics[width=0.17\textwidth]{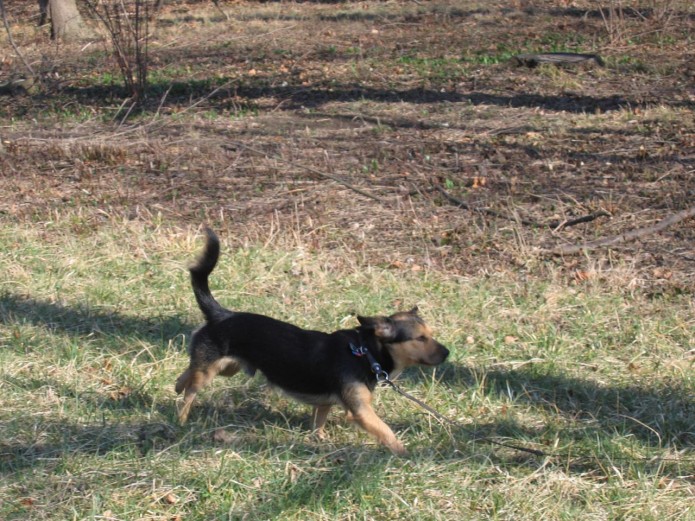}
 &\includegraphics[width=0.17\textwidth]{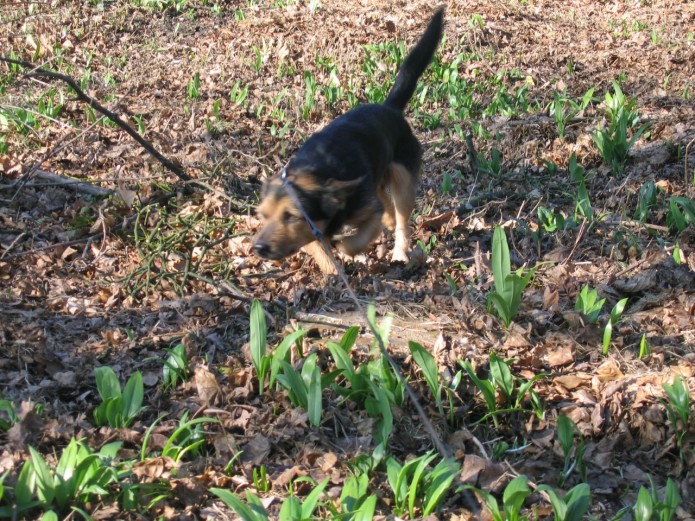} \\ \hline
\begin{tabular}[c]{@{}l@{}}Original \\ story\end{tabular} & \begin{tabular}[c]{@{}l@{}}The dog is\\  ready to go.\end{tabular} & \begin{tabular}[c]{@{}l@{}}He is playing \\ on the ground.\end{tabular}      \\ \hline
Terms                                                     & \begin{tabular}[c]{@{}l@{}}``Dog\_Noun'',\\  ``Motion\_Frame''\end{tabular}             & \begin{tabular}[c]{@{}l@{}}``Performers\_and\_roles\_Frame'',\\ ``Ground\_Noun''\end{tabular}                   \\ \hline\hline
\begin{tabular}[c]{@{}l@{}}Story \\ w/ CR\end{tabular}    & \begin{tabular}[c]{@{}l@{}}The dog is \\ ready to go.\end{tabular} & \begin{tabular}[c]{@{}l@{}}The dog is playing \\ on the ground.\end{tabular} \\ \hline 
Terms                                                     & \begin{tabular}[c]{@{}l@{}}``Dog\_Noun'',\\ ``Motion\_Frame''\end{tabular}             & \begin{tabular}[c]{@{}l@{}}``Dog\_Noun'',\\ ``Performers\_and\_roles\_Frame'',\\  ``Ground\_Noun''\end{tabular}           \\ \hline
\end{tabular}
}
\caption{Term extraction and coreference replacement (CR). Image
order is from left to right.}
\label{tab:term_example}
\end{table}

\paragraph{Repetition Penalty in Story Generation}
Instead of using a repetition penalty for predicted terms only, we use inter- and
intra-sentence repetition penalties with beam search to reduce redundancy in the
visual story~\cite{hsu2018using}. The word score at each beam search step is
\begin{align*}
\mbox{beam\_score}(x) = \log(p(x))-\alpha\cdot\mathds{1}(x{\in}S)\\-(\gamma/l)\cdot\mathds{1}(x{\in}R),
\end{align*}
where $\alpha$ and $\gamma$ are hyperparameters empirically set to 20 and
5. Set $S$ denotes words that appear in the current sentence, and set $R$
denotes words in previous sentences. The current story's length $l$
is a regularization term which reduces the penalty to prevent the story
generator from refusing to repeat words when generating grammatically
correct sentences as the length of the sentence or story increases.

\section{Experimental Setup}
In this paper, \system leverages the
object detection model to find reliable candidates for story casting, which
provides robust grounding.
We also use the image knowledge graphs such as Visual
Genome~\cite{krishnavisualgenome} to extract activities to correlate the roles
selected by the system.
FrameNet~\cite{baker1998berkeley} terms are used to present the semantic
concepts of selected roles and their extracted correlated activities for the
story plot, which later guides the model to compose the visual story from a
panoramic point of view.
We describe the experimental setup in detail in this section.

\subsection{Data Preparation}
Four datasets were used in this paper: Visual Genome, OpenIE, ROCStories Corpora,
and VIST Dataset.
VIST Dataset provides image-to-term materials for learning in Stage 1. For Stage 2, the object relations from Visual Genome or the term relations from OpenIE are the materials for Stage 2. 
ROCStories Corpora supplies a large quantity of pure textual stories for generation in Stage 3, and the VIST Dataset, the sole end-to-end visual storytelling dataset, is used to fine-tune the model. Note that when we here refer to all the available textual stories, these are from the VIST Dataset and the ROCStories Corpora. 

\paragraph{Visual Genome} 
Visual Genome contains labels for tasks in language and vision fields such as
object detection and object relation detection. 
Visual Genome has 108,077 images, 3.8 million object instances, and 2.3 million
relationships. We use it to pretrain the object detection model
used in Stage 1. In addition, we utilize the noun and verb relations provided by
the scene graph of Visual Genome to brainstorm activities which could link two
photos.   
Relations of nouns and verbs in each image have been labeled as \emph{(subject,
verb, object)}. 
The reasons we use verb relations are twofold. 
First, compared to unseen objects,
activities are better ingredients to add in visual stories for correlation
as they do less harm to grounding. Second, as most image-to-text datasets focus
on objects, stories generated based on them are relatively static. Hence adding
activities makes our stories vivid.  

\paragraph{OpenIE} 
OpenIE is an information extractor that finds relationships in 
sentences~\cite{pal2016demonyms,christensen2011analysis,saha2017bootstrapping,saha2018open}. 
Here we use its relations extracted from various corpora as another knowledge
base.\footnote{https://openie.allenai.org/}
Note that we only extract one-hop relations from OpenIE.

\paragraph{ROCStories Corpora} 
We use ROCStories, which contains 98,159 stories, to train the story
generator~\cite{mostafazadeh2016roc}. 
As the annotators of ROCStories were asked to write a five-sentence story given
a prompt, stories focus on a specific topic with strong logical inference.
Similar to VIST, we extract terms from sentences of stories which are used as
the input data of the story generation model.

\paragraph{VIST Dataset}
This is a collection of 20,211 human-written visual stories and 81,743 unique photos.
Each story in VIST contains five sentences and five corresponding photo images.
To train our image-to-term prediction model, we first extract terms from
sentences of stories in the VIST dataset as the gold labels.
Inspired by Semstyle~\cite{Alexander2018Semstyle}, we consider the key
components of the sentence to be its noun (object) and verb (action) terms;
therefore, given a sentence, we first detect nouns and verbs using a
part-of-speech tagger\footnote{SpaCy: https://spacy.io/}. Then, we extract verb
frames to replace original verbs using Open-SESAME~\cite{swayamdipta17open}, a
frame-semantic parser which automatically detects
FrameNet~\cite{baker1998berkeley} frames. For example, in
Table~\ref{tab:term_example}, the terms extracted from \textit{``The dog is ready to
go"} are ``Dog\_Noun" and ``Motion\_Frame".

As for the term-to-story generation model, the Transformer generator used in
Stage 3 is first trained on the ROCStories dataset with the textual stories and
their extracted terms, and then fine-tuned on both the ROCStories and VIST
datasets.

\subsection{Hyperparameter Configuration}
In all of our experiments, we used the same hyperparameters to train our model.
The hidden size of the term prediction and story generation models was set to
512. The head and layer number of the Transformer encoder were 2 and 4.
Both models
were trained with the Adam optimizer with an initial learning rate of 1e-3,
which decayed with the growth of training steps. 
During decoding, the beam size was set to 3 for both modules.

\section{Results and Discussion}
\begin{figure*}[t]
    \centering
    \includegraphics[width=\textwidth]{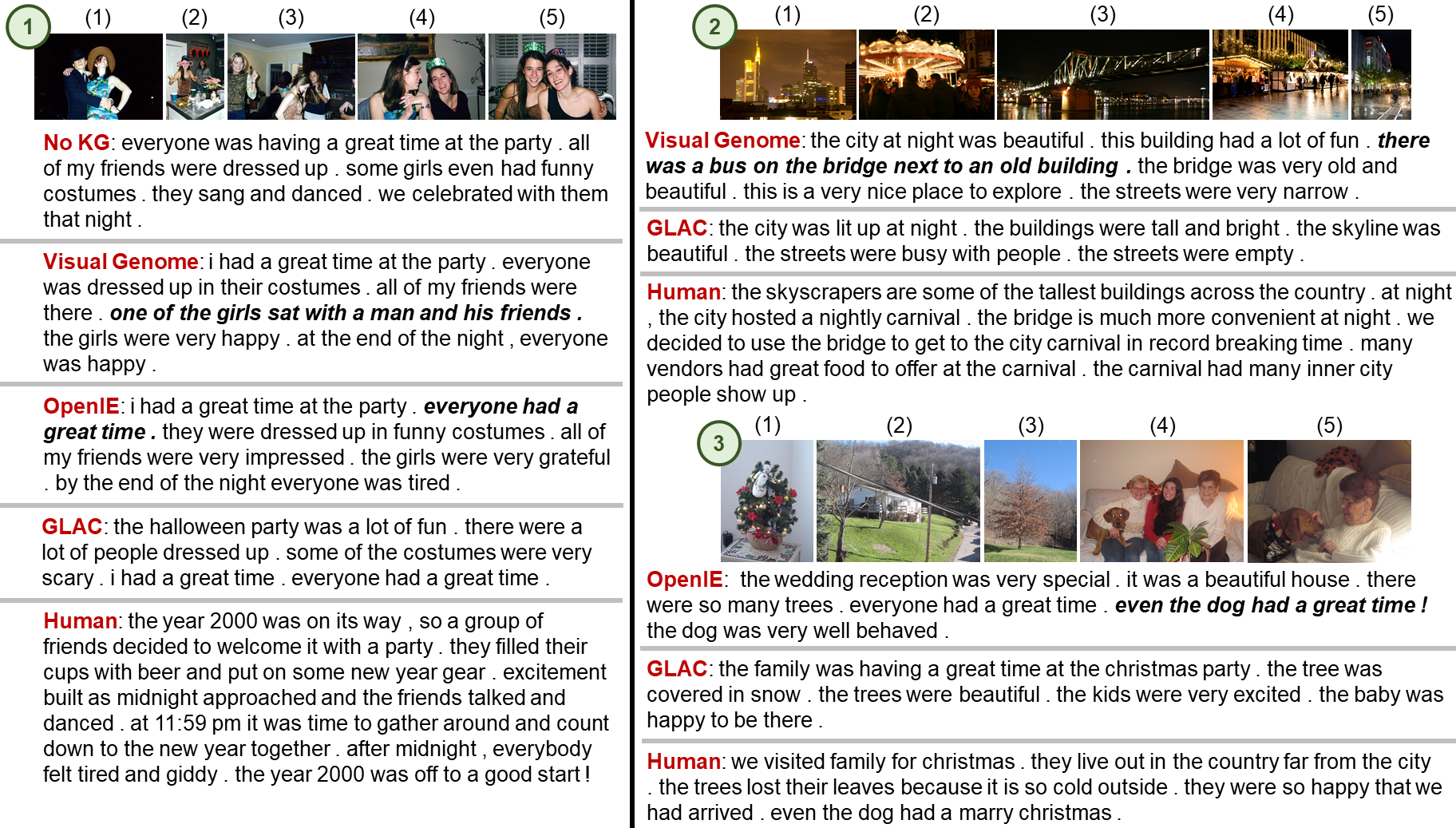}
	 \caption{Example stories generated by visual storytelling models. The knowledge-enriched sentences are highlighted.}
    \label{fig:case_study}
\end{figure*}

We conducted evaluations using both automatic
metrics and human ranking.
Automatic metrics were evaluated by shallow features such as the word coverage
or semantics for the quality stories~\cite{wang2018nometics}, whereas human
evaluation provided solid results due to their complete comprehension. 
For human evaluations, we adopted the ranking of stories from all models and the
stories written by humans. All human evaluations were conducted on the Amazon Mechanical Turk; five different workers were assigned for each comparison.
Through this direct comparison, we were able to provide a baseline for turkers,
yielding reliable results.

\subsection{Human Evaluation}

\begin{table}[t]
\resizebox{1\columnwidth}{!}{
\begin{tabular}{@{}lrrrrr@{}}
\multicolumn{6}{c}{Human Evaluation (Story Displayed \textbf{with} Images)} \\ \midrule
\textbf{} & \textbf{\begin{tabular}[c]{@{}r@{}}GLAC\\ \cite{Kim2018GLACNG}\end{tabular}} & \textbf{No KG} & \textbf{OpenIE} & \textbf{\begin{tabular}[c]{@{}r@{}}Visual\\ Genome\end{tabular}} & \textbf{Human} \\ \midrule
\textbf{\begin{tabular}[c]{@{}l@{}}Avg. Rank\\ (1 to 5)\end{tabular}} & 3.053 & 3.152 & \textbf{2.975*} & \textbf{2.975*} & 2.846 \\ \bottomrule
\end{tabular}
}
\caption{Direct comparison evaluation of \system model. Numbers indicate average rank given to stories (from 1 to 5, lower is better.) Stories generated by \system using either OpenIE or Visual Genome are on average ranked significantly better (lower) than that of GLAC (unpaired t-test, $p<0.05$, N=2500).}
\label{tab:comparison}
\end{table}


\paragraph{Ranking Stories with Images}
We conducted a human evaluation using crowd workers recruited from Amazon
Mechanical Turk.
Each task displayed one sequence of photos and several generated stories based on
the sequence.
Five workers were recruited for each task to rank the story quality, ``from the
Best Story to the Worst Story.''
The compensation was \$0.10 per task.   
Four models were included in this evaluation: 
{\em (i)} \system using OpenIE as the knowledge graph,
{\em (ii)} \system using VisualGenome as the knowledge graph,
{\em (iii)} \system without any knowledge graphs, and {\em (iv)} GLAC~\cite{Kim2018GLACNG},
the current state-of-the-art visual storytelling model. 
Note that in the first two models, only the knowledge graph for exploring inter-image relations in Stage 2 is different, while Stage 1 and Stage 3 remain identical.
As we also included {\em (v)} human-written stories for comparison, the ranking
number is from 1 (the best) to 5 (the worst).
Table~\ref{tab:comparison} shows the evaluation results, showing that \system
benefits from the enrichment of knowledge. 
Stories generated by \system using either OpenIE or Visual Genome are on
average ranked significantly better (lower) than that of GLAC (unpaired t-test,
$p<0.05$, N = 2500), 
whereas stories generated by \system without using any knowledge graphs ranked
significantly worst (higher) (unpaired t-test, $p<0.01$, N = 2500).

\paragraph{Ranking Stories without Images}
\begin{table}[t]
\resizebox{1\columnwidth}{!}{
\begin{tabular}{@{}lrrrrr@{}}
\multicolumn{6}{c}{Human Evaluation (Story Displayed \textbf{without} Images)} \\ \midrule
\textbf{} & \textbf{\begin{tabular}[c]{@{}r@{}}GLAC\\ \cite{Kim2018GLACNG}\end{tabular}} & \textbf{No KG} & \textbf{OpenIE} & \textbf{\begin{tabular}[c]{@{}r@{}}Visual\\ Genome\end{tabular}} & \textbf{Human} \\ \midrule
\textbf{\begin{tabular}[c]{@{}l@{}}Avg. Rank\\ (1 to 5)\end{tabular}} & 3.054 & 3.285 & \textbf{3.049} & \textbf{2.990} & 2.621 \\ \bottomrule
\end{tabular}
}
\caption{Direct comparison evaluation without given photos. Numbers indicate average rank given to stories (from 1 to 5, lower is better).}
\label{tab:comparison-no-img}
\end{table}

We also explored the relation between images and the generated stories
in visual storytelling.
To determine whether the generated stories are well matched to the photos, we
removed the photos and conducted another overall story comparison. Results in
Table~\ref{tab:comparison-no-img} show that when
photos were provided in human evaluation, visual stories from the \system were
ranked higher than those from the state-of-the-art GLAC (2.97
vs. 3.05, significant with p-value $<$ 0.05); they become comparable (3.04 vs.
3.05) when the photos were not provided. The ranking of visual stories from the
state-of-the-art model remains the same. This
suggests that visual stories from \system better fit the photos, where
the images and texts support each other. 
The third story in   
Figure~\ref{fig:case_study} illustrates an example. The visual story from
the state-of-the-art GLAC clearly is not grounded: there is no snow in
the second photo, no kids in the fourth, and no baby in the fifth, though
only considering the text it is a reasonable Christmas story. In contrast,
given the photos, the proposed \system generates an obviously better
story. This example also illustrates that the given images are both hints and
constraints in visual storytelling, which makes it a more challenging task than
general story generation. Besides, even when photos are not provided, the
average rank of visual stories from \system is comparable to that from the state-of-the-art
model. This demonstrates that our stories are both closely aligned to the
images and also stand without the images, which we attribute to the use
of the additional Virtual Genome dataset and ROCStories Corpora for the model
learning.

\paragraph{The Effect of Using Knowledge Graphs}

\begin{figure}[htbp]
    \centering
    \includegraphics[width=\columnwidth]{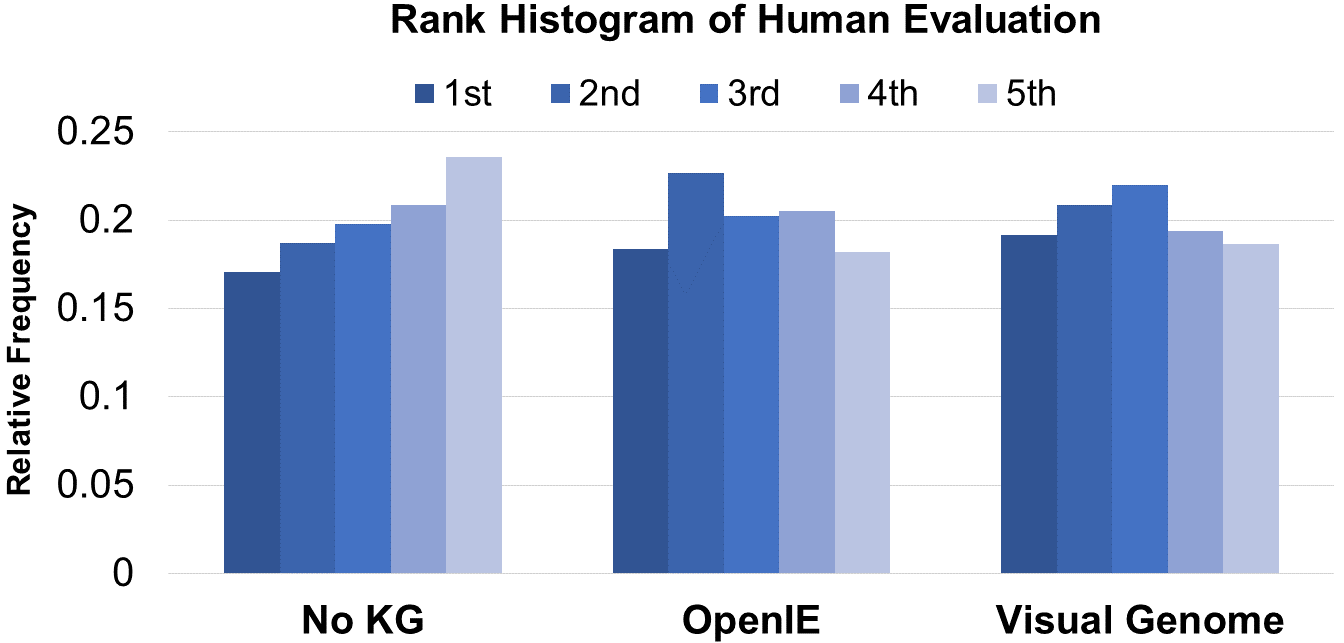}
	 \caption{The distribution of the ranks received by each model. From the darkest blue to the lightest blue is the best rank to the worst.}
    \label{fig:histogram}
\end{figure}

We also investigate the effect of using a knowledge graph on story
quality.
Figure~\ref{fig:histogram} shows the distribution of ranks assigned to
models with and without the enrichment of the knowledge graph. Overall we note a clearly
increasing number of stories ranked as the first, second, and third
place when considering the knowledge graph, which confirms its
benefit to the quality of stories.
Specifically, stories from the model with the Visual Genome receive more first
ranks and also more worst ranks than those from the model with OpenIE. In
other words, compared to stories enriched by Virtual Genome, those from OpenIE
are relatively moderate. This may result from the existence of the two-hop
relations in Visual Genome, whereas OpenIE contains only one-hop relations. As a
result, the model with Visual Genome tends to be more imaginative and
unrestrained, leading to stories that are either extremely good or extremely
bad. 

\paragraph{Observing Automatic Metrics}

As prior work has clearly shown that the classic automatic evaluation metrics are
weak or even negative quality indicators for visual 
storytelling~\cite{hsu-etal-2019-visual,wang2018nometics}, we do not use these for
evaluation.
However, they still have something to say about token-based textual analysis.
Interestingly, we observe that the use of knowledge graphs increases the scores
in precision-based metrics (i.e., BLUE1, BLEU2, and CIDEr) with the cost of slightly
reduced scores in recall-based metrics (Table~\ref{tab:auto_eval}).

\begin{table}[htbp]
\resizebox{1\columnwidth}{!}{
\begin{tabular}{@{}lrrrrr@{}}
\toprule
 & \textbf{BLEU1} & \textbf{BLEU4} & \textbf{METEOR} & \textbf{ROUGE} & \textbf{CIDEr} \\ \midrule
\textbf{GLAC~\cite{Kim2018GLACNG}} & .374 & .050 & \textbf{.297} & \textbf{.250} & .049 \\ \midrule
\textbf{No KG} & .376 & .050 & .295 & .237 & .055 \\
\textbf{OpenIE} & .446 & \textbf{.056} & \textbf{.297} & .241 & .094 \\
\textbf{Visual Genome} & \textbf{.451} & \textbf{.056} & .296 & .241 & \textbf{.096} \\ \bottomrule
\end{tabular}
}
\caption{Automatic evaluation results. Prior work has clearly shown that the classic automatic evaluation metrics are
weak or even negative quality indicators for visual 
storytelling~\cite{hsu-etal-2019-visual,wang2018nometics}.}
\label{tab:auto_eval}
\end{table}

\subsection{Discussion}

In this section, we discuss some of our observations throughout the
experiments.

\paragraph{A New Relation Bridges Two Images}
We observe that often the newly added relation creates bridges between two
images.
For example, for the first photo sequence in Figure~\ref{fig:case_study},
\system with Visual Genome distills \textit{friends} from the third image and
\textit{girls} from the fourth image.
Then the system connects the two terms with the two-hop relation \emph{(friends, posture,
man)} and \emph{(man, posture,
girls)}.
Similarly, \system with OpenIE links \emph{time} from the first image and \emph{everyone} from
the second image with the relation \emph{(time, cause to experience, everyone)}. 
We also find that after adding KG terms to the term set, the entire generated
story is different from the No-KG stories, even though the terms for the
original images are not changed. This shows that \system indeed considers all terms together when generating stories, and hence added terms influence its context.

\paragraph{Low-ranked Human Generated Stories}

From Table ~\ref{tab:comparison} and Table ~\ref{tab:comparison-no-img} we notice that human generated stories are not ranked prominently better comparing the the ranked given to machine generated stories. In Table ~\ref{tab:comparison} it is 2.85 compare to 2.97 and in Table ~\ref{tab:comparison-no-img} it is 2.62 compare to 2.99. In addition to the good quality of our generated stories, we found that this is also because in the evaluation we allow crowd workers to rank stories directly by comparison without explicitly giving pre-defined rubrics. Since people may have different definition for a good story, the machine-generated stories were not always ranked lower than human-written ones. For example, some people may value an interesting plot more than a focused or detailed description. 

\paragraph{Repetition Penalty Reduces Repetition}

We observe that there are repetitions in GLAC's stories. For example, in story ``\emph{the new office is a very interesting place. the kids love the toy. they also have a lot of books. they have a nice collection of books. they even have a stuffed animal}'' and ``\emph{the kids were having a great time at the party. they had a lot of fun. they had a lot of fun. they had a lot of fun. they had a great time.}'', 
the pattern ``\emph{they have}'' and ``\emph{they had a lot of fun}'' both respectively appear three times, which is generally less human and not ideal.
We find that this phenomenon is reduced by the repetition penalty in
\system.

\paragraph{When Nonexistent Entities Are Added}
Adding terms that refer to entities or people that are not actually in the
image is a potential risk when enriching stories with information from the KG. 
We find this problem to be especially severe for two-hop relations.
The second photo sequence in Figure~\ref{fig:case_study} is an example in which the KG fails to help.
After \emph{building} is distilled from the second image and \emph{bridge} is distilled
from the third image, \system selects the link for enrichment: $\mathit{building}
\xrightarrow{\text{obscurity}} \mathit{bus} \xrightarrow{\text{travel}} \mathit{bridge}$.
While this link itself as well as the generated sentence ``\emph{there was a bus on
the bridge next to an old building}'' are both quite valid, it does not align
to the images as the bus, the building, and the bridge are not in the
described relative positions. 
In this case, the model generates unsatisfactory stories.

\begin{figure*}[htbp]
    \centering
    \includegraphics[width=1\textwidth]{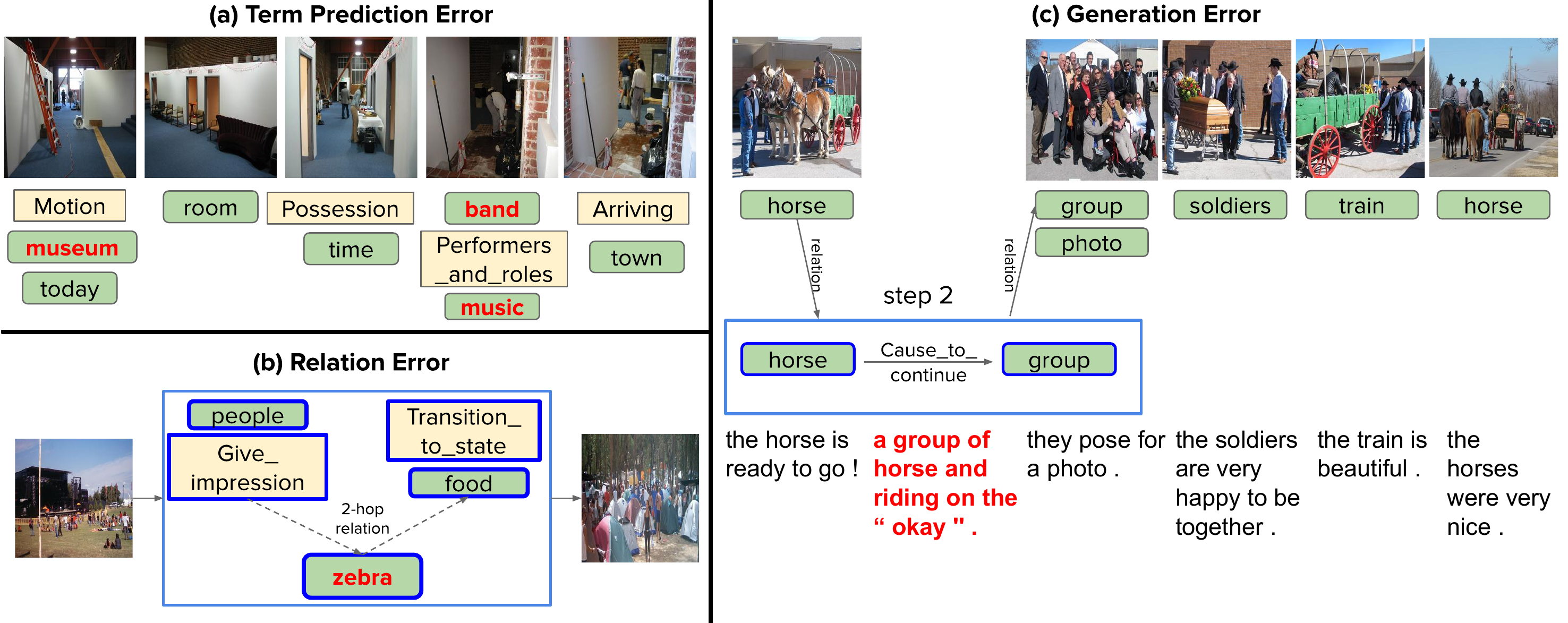}
	 \caption{Error analysis examples. Green cells are noun terms. Yellow cells are verb terms. Blue border cells are terms enriched by step 2. Red text cells are the identified errors. (a) shows a term prediction error where the model predicted unrelated terms museum, band and music for pictures about construction; in (b), the model selected a 2-hop relation which contained an unrelated object zebra from the knowledge graph to add in, causing a relation error; in (c) a generation error was identified due to the incomprehensible sentence marked in red.}
    \label{fig:error_example}
\end{figure*}

\paragraph{Error Analysis}

To understand the behavior of \system, we conducted a small-scale error analysis on 50 stories. We randomly selected stories for examination from 1,999 stories with their corresponding photo sequences, and kept those stories containing errors until we reached the number 50. Then these 50 stories were manually examined.
Each of the 50 stories exhibit one of the following errors: term prediction error, relation error, and generation error.
Figure ~\ref{fig:error_example} shows the example of each error type.
The following describes a distribution of the errors from our analysis:

\begin{enumerate}
    \item \textbf{Term Prediction Error (Stage 1): 48\%.}
    The term prediction error occurs when the story or terms are redundant or unrelated to the images content regardless the additional sentence. 
    The term that were unrelated to the images caused the model to generate irrelevant story, and thus, deteriorate the overall performance. 
    \item \textbf{Relation Error (Stage 2): 26\%.}
    The relation error means the extracted relations are redundant with existing terms or does not match with visual content in the images. 
    Bounding the added relations with images may further mitigate their impact.
    \item \textbf{Generation Error (Stage 3): 26\%.}
    Generation error refers to the low-quality generated sentence under the condition that the given terms are valid. 
    This may result from some rarely seen combination of terms for which the generation model does not know how to describe them. Collecting a large quantity of diverse stories for learning could be a solution.
    
\end{enumerate}
In addition to random selection, we also identified 12 additional low-ranked KG-stories and their errors to find the cause of low ranking. We checked stories that were ranked 4 and 5 in the previous evaluation. After the analysis, the error distribution was 33\%, 17\%, and 50\% for term prediction error, relation error, and generation error, respectively. Compared to the errors in the randomly selected stories, the decrease percentage of the grounding error and the increase percentage of the sentence generation error suggest that people tend to give relatively low rating when seeing simple, easily detected errors in story sentences.  

\section{Conclusion}
In this paper, we seek to compose high-quality, visual stories that are enriched by the
knowledge graph.
We propose \system, a novel three-stage visual storytelling model which 
leverages additional non-end-to-end data.
In the generation process we address 
positional information, anaphors, and repetitions.
Human evaluation shows that \system outperforms the state of the art in both
automatic evaluation and human evaluation for direct comparison.
In addition, we show that even when evaluating without the corresponding images, the
generated stories are still better than those from the state-of-the-art model,
which demonstrates the effectiveness of improving coherence by the knowledge
graph and learning from additional uni-modal data.
Given these encouraging results, composing arbitrary-length stories from the same
input hints is our next goal.

\section{Acknowledgements}
This research is partially supported by Ministry of Science and Technology, Taiwan under the project contract 108-2221-E-001-012-MY3, 108-2634-F-001-004-, and the Seed Grant (2019) from the College of Information Sciences and Technology (IST), Pennsylvania State University.

\bibliographystyle{aaai}
\bibliography{aaai.bib}

\end{document}